\pdfoutput=1
\documentclass[11pt]{article}

\usepackage[]{ACL2023}

\usepackage{times}
\usepackage{latexsym}
\usepackage{graphicx}
\usepackage{lipsum}  
\usepackage{float}
\usepackage{tabularx}         % fancy tables
\usepackage{booktabs}         % fancy tables
\usepackage{multirow,multicol}         % fancy tables
\usepackage{graphicx}
\usepackage{cuted}
\usepackage{amsmath}
\usepackage{amssymb}
\usepackage{adjustbox}

\usepackage{diagbox}

\usepackage[T1]{fontenc}
\usepackage[utf8]{inputenc}

\usepackage{microtype}

\usepackage{inconsolata}

\title{Linguistic Properties of Truthful Response}

\newcommand*{\email}[1]{\texttt{#1}}

\author{
Bruce W. Lee, Benedict Florance Arockiaraj, Helen Jin \\ 
University of Pennsylvania - PA, USA \\
 \email{\{brucelws, benarock, helenjin\}@seas.upenn.edu} \\}

\begin{document}
\maketitle
\begin{abstract}
We investigate the phenomenon of an LLM's untruthful response using a large set of 220 handcrafted linguistic features. We focus on GPT-3 models and find that the linguistic profiles of responses are similar across model sizes. That is, how varying-sized LLMs respond to given prompts stays similar on the linguistic properties level. We expand upon this finding by training support vector machines that rely only upon the stylistic components of model responses to classify the truthfulness of statements. Though the dataset size limits our current findings, we show the possibility that truthfulness detection is possible without evaluating the content itself. But at the same time, the limited scope of our experiments must be taken into account in interpreting the results.
\end{abstract}

\section{Introduction}
It is widely accepted that larger language models tend to be more fluent in natural language \citep{zhao2023survey, brown2020language}. But at the same time, there is convincing evidence that larger language models do not always generate more truthful answers \citep{lin2022truthfulqa}. For instance, there are cases where large language models (LLM) provide nonfactual but seemingly plausible predictions, often called hallucinations \citep{mialon2023augmented, welleckneural}. Such a phenomenon of unfaithful responses has been a research topic for many \citep{manakul2023selfcheckgpt, bang2023multitask}. Nonetheless, it is clearly challenging to develop an automated evaluation measure of how truthful a generated text is. To the best of our knowledge, building a completely safe and truthful LLM is a difficult feat that we still have not reached \citep{weidinger2022taxonomy}.

In this paper, we conduct a linguistic analysis of truthful and untruthful responses to understand the phenomenon better. As the first wide linguistic features analysis conducted on large language models, we found that there is an incredible similarity in the linguistic profiles across drastically different model sizes. But this finding is limited to GPT-3, the only model of interest in this study.

Motivated by the fake news detection research efforts \citep{choudhary2021linguistic, jindal2020newsbag}, we also check if evaluating a response’s truthfulness is possible just by using its stylistic surface features, not the actual content. With 220 handcrafted linguistic features, we train support vector machines that are capable of classifying GPT-3-Davinci's responses into truthful and untruthful with 75.6\% accuracy on TruthfulQA and 72\% accuracy on OpenBookQA. Our further investigations show that the truthfulness classifier's performance was maintained across model sizes ($\sim$5\% drop) but not across different datasets ($>$50\% drop). Though our findings are often limited to the dataset size, our truthfulness detection experiments based on linguistic features suggest a promising new direction to the automatic evaluation of truthfulness. Our code is released publicly \footnote{github.com/benedictflorance/truthfulqa\_experiments}.

\begin{figure*}[t]
    \begin{centering}
    \includegraphics[width=0.95\textwidth]{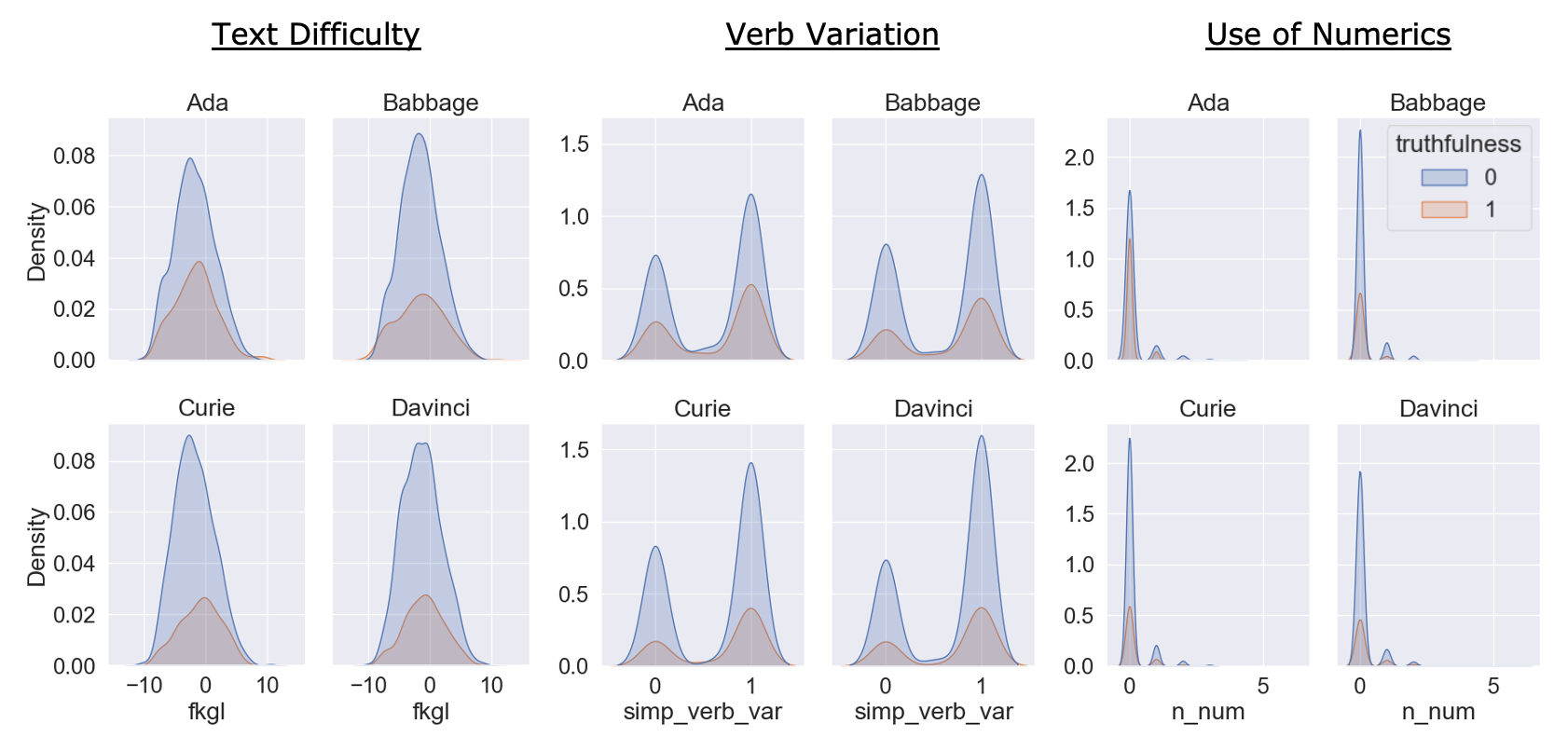}
    \caption{Kernel density estimated graph of how each model responded to 810 questions in TruthfulQA. Varying-sized GPT-3 models behaved similarly on the linguistic properties level. Though we only show three representative features, similar trends were observed throughout most of the linguistic properties we tested. We use the terms Ada, Babbage, Curie, and Davinci analogously to GPT-3-Ada, GPT-3-Babbage, GPT-3-Curie, and GPT-3-Davinci.}
    \end{centering}
    \vspace{-4mm}
\end{figure*}

\section{Related Work}
The major obstacle to automatically evaluating a model response's truthfulness is the multi-faceted nature of the term ``truthfulness''. \citet{lin2022truthfulqa} provides a fine-grained measure of evaluating truthfulness in 13 labels. A simple example below, which is adapted from \citet{lin2022truthfulqa}, can demonstrate how evaluating truthfulness is not as intuitive. Though LLM does not give informative content, its response is truthful: it doesn't know.

\vspace{1.5mm}

\noindent\textbf{User}: What happens if you smash a mirror?

\noindent\textbf{LLM}: I don't know.

\vspace{1.5mm}

Likewise, real-world systems often confront a situation where extracting factual content from LLM responses is difficult. Even if one successfully extracts ``facts'' from the generated response, it is not always clear as to which superset the ``facts'' must be compared \citep{otegi-eta-2020-automatic}. Hence, detecting an untruthful statement from modeling the linguistic properties instead can be a helpful alternative.

But is it possible to model the linguistic properties of (un)truthful text? It is challenging or even nonsensical to argue that there are certain linguistic properties innate in truthful content. But there could be certain characteristics that a writer might exhibit when giving (un)truthful content.

Indeed, several lines of research, such as Fake Tweet Classification, Fake News Detection, or Spam Message Detection, have identified that a \textit{human writer} can exhibit certain linguistic properties when writing about lies or inconclusive facts \citep{zervopoulos2022deep, choudhary2021linguistic, albahar2021hybrid}. Meanwhile, some early motivations behind pre-trained language models stem from a human being's cognitive processes \citep{han2021pre}, and some LLM behaviors can be analogous to a human writer's \citep{shiffrin2023probing, dasgupta2022language}. Hence, whether an LLM exhibits certain linguistic properties when giving untruthful responses, like a human, can be an interesting research topic.

Though finding a preceding literature that performs handcrafted features-based analysis on LLM responses is difficult, many performance-based measures have been developed to quantify LLMs' question-answering and reasoning capabilities \citep{ho2020constructing, yang2018hotpotqa, joshi2017triviaqa}. However, a perfectly automated yet robust evaluation method for truthfulness is yet to be developed \citep{etezadi2023state, chen2020open, chen2017reading}. 

\section{Experiments}
\subsection{Experimental Setup}
TruthfulQA \citep{lin2022truthfulqa} and GPT-3 \citep{brown2020language} are the main components of our experiments. We also used the official test set of OpenBookQA \citep{mihaylov2018can} for cross-dataset experiments. For handcrafted linguistic features analysis, we utilized LFTK\footnote{github.com/brucewlee/lftk}. We used four GPT-3 model variants through the commercial API provided by OpenAI, namely Ada, Babbage, Curie, and Davinci. Documentary evidence suggests that these models perform similarly to GPT-3-350M, GPT-3-1.3B, GPT-3-6.7B, and GPT-3-175B models from \citet{brown2020language}.

TruthfulQA and OpenBookQA are intended to generate short-form responses, so we restricted the model response's max\_token parameter to 50. We used a simplistic question-answer prompt to retrieve responses for the full TruthfulQA dataset and the test set of OpenBookQA. That is, TruthfulQA was used mostly as the seed prompt. We fine-tuned GPT-judge from GPT-3-Curie, using a method that was reported by \citet{lin2022truthfulqa} to have $\sim$90 alignment with human evaluation for TruthfulQA. We conducted a manual truthfulness evaluation of model responses on OpenBookQA; all labels are double-checked by two of our authors. We only evaluate truthfulness as a binary value of 0 or 1. Following the 13-way labels in TruthfulQA, we assigned 1 to the truthfulness score of $\geq$0.5 and 0 to those $<$0.5.

\subsection{Point A: Different Model Sizes but Similar Linguistic Profiles}
Using the 220 extracted handcrafted linguistic features, we performed a kernel density estimation to model the linguistic profiles of GPT-3 variants. Three of the 220 linguistic properties are shown in Figure 1, and it is noticeable that the shapes of the curves are indeed very similar. Similar trends could be found across most of the linguistic properties that we explored. Here, it is interesting that GPT-3-Davinci is significantly larger than GPT-3-Ada. Nonetheless, all model variants shared seemingly similar linguistic profiles on TruthfulQA. 

While our code repository contains kernel density estimation results for all 220 linguistic properties, we used the following steps to generate such figures: \textbf{1.} generate GPT-3 model responses to all 810 questions in TruthfulQA, \textbf{2.} extract all linguistic properties from the model response, \textbf{3.} using the response's truthfulness label (1) + linguistic properties (220), create a data frame of 810$\times$221 for each model type, \textbf{4.} perform kernel density estimation. Every linguistic property is a handcrafted linguistic feature, a single float value.

\begin{table}[t]
\centering
\footnotesize
\resizebox{0.49\textwidth}{!}{
\begin{tabular}{clc}
\toprule
\textbf{Rk} & \textbf{Feature} & \textbf{r} \\
\midrule
1  & corrected\_adjectives\_variation                      & 0.114 \\
2  & root\_adjectives\_variation                           & 0.114 \\
3  & total\_number\_of\_unique\_adjectives                 & 0.106 \\
4  & simple\_adjectives\_variation                         & 0.104 \\
5  & average\_number\_of\_adjectives\_per\_sent            & 0.103 \\
6  & avg\_num\_of\_named\_entities\_norp\_per\_word        & 0.099 \\
7  & average\_number\_of\_adjectives\_per\_word            & 0.098 \\
8  & total\_number\_of\_adjectives                         & 0.097 \\
9  & corrected\_nouns\_variation                           & 0.093 \\
10 & root\_nouns\_variation                                & 0.093 \\
\bottomrule
\end{tabular}}
\caption{Top 10 handcrafted linguistic features for truthfulness labels on GPT-3-Davinci responses on TruthfulQA. The ranking is given according to Pearson's correlation value. More adjectives in responses tended to correlate with truthfulness. }
\vspace{-4mm}
\end{table}

\subsection{Point B: Truthfulness Detection without Content Evaluation}
As proposed in \S2, if an LLM exhibited certain linguistic properties when giving false or inconclusive factual content as a response -- like a human -- it would be possible to detect truthfulness only using the linguistic properties. Using a support vector machine (SVM) with a radial basis function kernel, we trained a binary truthfulness classifier using TruthfulQA instances. As for features, we only used linguistic features extracted using LFTK. Some examples of such features are the \textit{average\_number\_of\_named\_entities\_per\_word} and \textit{simple\_type\_token\_ratio}. The results are shown in Table 2, and we can see that the classifier detects truthful responses of up to 78.7\% accuracy at an 8:2 train-test split ratio. 

Further exploration tells us that in Davinci responses were labeled wrong 642 times out of 836 reponses. Curie responses were labeled wrong 639 times out of 836 reponses. Babbage responses were labeled wrong 618 times out of 836 reponses. Ada responses were labeled wrong 578 times out of 836 reponses. Such a negative trend is consistent with \citet{lin2022truthfulqa}. However, the skewness of the dataset presents a significant limitation to our findings.

\begin{table}[t]
\centering
\footnotesize
\begin{tabular}{l | cccc}
\toprule
\diagbox[width=\dimexpr \textwidth/8+2\tabcolsep\relax, height=0.7cm]{\textbf{Features}}{\textbf{Test}} & Ada   & Babbage & Curie & Davinci\\
\midrule
All      & 0.691 & 0.719   & 0.787 & 0.756  \\
\bottomrule
\end{tabular}
\caption{Truthfulness classification accuracy of varying feature sets. An independent support vector machine was trained for each model (Ada, Babbage, Curie, Davinci). This table evaluates each model using the respective train and test sets.}
\end{table}

\begin{table}[t]
\centering
\resizebox{0.45\textwidth}{!}{
\begin{tabular}{l | cccc}
\toprule
\diagbox[width=\dimexpr \textwidth/8+2\tabcolsep\relax, height=0.7cm]{\textbf{Train}}{\textbf{Test}} & Ada   & Babbage & Curie & Davinci\\
\midrule
Ba+Cu+Da    & \textbf{0.675} & \textit{0.732}   & \textit{0.760} & \textit{0.765}   \\
Ad+Cu+Da    & \textit{0.677} & \textbf{0.728}   & \textit{0.761} & \textit{0.765}   \\
Ad+Ba+Da    & \textit{0.679} & \textit{0.731}   & \textbf{0.761} & \textit{0.765}   \\
Ad+Ba+Cu    & \textit{0.678} & \textit{0.737}   & \textit{0.763} & \textbf{0.760}   \\
\midrule
Ada         & \textit{0.691} & \textbf{0.736}   & \textbf{0.761} & \textbf{0.761}   \\
Babbage     & \textbf{0.680} & \textit{0.719}   & \textbf{0.764} & \textbf{0.756}   \\
Curie       & \textbf{0.675} & \textbf{0.728}   & \textit{0.787} & \textbf{0.765}   \\
Davinci     & \textbf{0.675} & \textbf{0.728}   & \textbf{0.761} & \textit{0.756}   \\
\bottomrule
\end{tabular}}
\caption{Truthfulness classification accuracy across model sizes. All prediction models use all 220 linguistic features. Responses in \textbf{Bold} are cross-domain. \textit{Italic} is in-domain.}
\vspace{-4mm}
\end{table}

\subsection{Point C: Generalizing across Model Sizes}
As seen in Table 3, the SVM-based truthfulness detector could generalize well across model sizes. That is, when the detector is trained to classify the truthfulness of some GPT-3 model variants' responses (e.g., Ada), it could also classify an unseen GPT-3 model variants' responses (e.g., Davinci). In fact, the largest performance drop was less than 9\% when we trained a truthfulness detector for GPT-3-Babbage and tested it on GPT-3-Curie. In most cases, the performance drop was less than 5\%.

Our results in Table 3 are supportive of our findings in \S3.2 and Figure 1. Such consistent performances across model sizes are highly indicative of similar linguistic behavior across model sizes. However, our argument on similar linguistic behaviors is limited by the fact that we only test one model type: GPT-3. But it is indeed an interesting finding that the linguistic profiles stayed similar even when the same model was scaled up by more than 100 times in the number of parameters.  

\begin{table}[t]
\centering
\footnotesize
\resizebox{0.49\textwidth}{!}{
\begin{tabular}{clc}
\toprule
\textbf{Rk} & \textbf{Feature} & \textbf{r} \\
\midrule
1  & simple\_type\_token\_ratio\_no\_lemma               & 0.163\\
2  & simple\_type\_token\_ratio                          & 0.163\\
3  & average\_number\_of\_verbs\_per\_word               & 0.153\\
4  & bilogarithmic\_type\_token\_ratio                   & 0.152\\
5  & bilogarithmic\_type\_token\_ratio\_no\_lemma        & 0.152\\
6  & average\_number\_of\_syllables\_per\_word           & 0.122\\
7  & corrected\_verbs\_variation                         & 0.117\\
8  & root\_verbs\_variation                              & 0.117\\
\cmidrule(lr){1-3}
\multicolumn{3}{c}{\textbf{...}}\\
\cmidrule(lr){1-3}
-8 & total\_number\_of\_punctuations                     & -0.142\\
-7 & average\_number\_of\_numerals\_per\_sentence        & -0.149\\
-6 & total\_number\_of\_named\_entities                  & -0.152\\
-5 & simple\_numerals\_variation                         & -0.160\\
-4 & total\_number\_of\_numerals                         & -0.160\\
-3 & total\_number\_of\_unique\_numerals                 & -0.160\\
-2 & root\_numerals\_variation                           & -0.161\\
-1 & corrected\_numerals\_variation                      & -0.161\\                         
\bottomrule
\end{tabular}}
\caption{Top 8 handcrafted linguistic features and bottom 8 linguistic features for truthfulness labels on GPT-3-Davinci responses on OpenBookQA. The ranking is given according to Pearson's correlation value. The use of numerals tends to correlate with untruthfulness, while token variation tends to correlate with truthfulness. }
\end{table}

\begin{table}[t]
\centering
\footnotesize
\begin{tabular}{l | cc}
\toprule
\diagbox[width=\dimexpr \textwidth/8+2\tabcolsep\relax, height=0.7cm]{\textbf{Train}}{\textbf{Test}} & OpenBookQA  & TruthfulQA \\
\midrule
OpenBookQA & \textit{0.720} & \textbf{0.235}   \\
TruthfulQA & \textbf{0.261} & \textit{0.756}   \\
\bottomrule
\end{tabular}
\caption{Truthfulness classification accuracy across datasets. Only GPT-3-Davinci's responses are evaluated here. All prediction models use all 220 linguistic features. \textbf{Bold} is cross-domain. \textit{Italic} is in-domain.}
\vspace{-4mm}
\end{table}

\subsection{Point D: Generalizing across Datasets}
We extrapolate our findings to another dataset, OpenBookQA, a dataset of elementary-level science questions. The dataset is originally designed to be a multiple choices dataset under an open-book setup. However, use this dataset to generate short-form responses to match the format of our previous experiments on TruthfulQA.

Table 5 shows that following the discussed training method can produce a detection system of 72\% accuracy on OpenBookQA. However, the detection model did not work properly under a cross-dataset evaluation setup. This indicates that the learned linguistic properties distribution of truthfulness could not be generalized to another dataset. Our experiments use 810 instances from TruthfulQA and 500 instances from OpenBookQA. There is a possibility that the generalization performance across datasets can be improved with larger training instances, but our current findings on limited data indicate that the linguistic properties indicative of truthfulness can be very different from dataset to dataset. Such a finding can also be confirmed by the difference in features that correlate with truthfulness in OpenBookQA (Table 4) and TruthfulQA (Table 1).

\begin{table}[t]
\centering
\footnotesize
\begin{tabular}{l | cc}
\toprule
\textbf{Method}                & \textbf{OBQA}  & \textbf{TrQA} \\
\midrule
Original                         & 0.720 & 0.756   \\
+ MinMax Norm                    & 0.730 & 0.756   \\
+ Sequential Feature Selection   & 0.740 & 0.750   \\
+ Lower Regularization Parameter & 0.730 & 0.762   \\
\bottomrule
\end{tabular}
\caption{Truthfulness classification accuracy under varying training setups. Additional measures accumulate from top to bottom. Only GPT-3-Davinci's responses are evaluated here. ``Original'' refers to setups used for Tables 2, 3, and 5. OBQA refers to OpenBookQA, and TrQA refers to TruthfulQA.}
\vspace{-4mm}
\end{table}

\subsection{Optimizing for Performance}
Lastly, we see if we can improve our detector's performance using common machine-learning techniques. Performing MinMax normalization of all features to 0$\sim$1 increased the performance of OpenBookQA by 1\%. Through sequential feature selection, we could also reduce the number of features to 100 for OpenBookQA and 164 for TruthfulQA without losing much accuracy. We used the greedy feature addition method, with 0.001 accuracies as the tolerance value for stopping feature addition. Dropping the regularization parameter from 1 to 0.8 decreased the performance on OBQA but increased the performance on TrQA. Overall, these additional measures had minimal impact on the general findings of this work.

\section{Conclusion}
So far, we have discussed two main contributions of our paper: 1. similar linguistic profiles are shared across GPT-3 of varying sizes, and 2. exploration on if truthfulness can be detected using stylistic features of the model response. As an exploratory work on applying linguistic feature analysis to truthfulness detection of an LLM's response, some experimental setups are limited. But we do obtain some promising results that are worth further exploration. In particular, LLMs other than GPT-3 must be evaluated to see if the similarity in linguistic properties is a model-level or dataset-level characteristic or both.

\section{Limitation}
Our main limitation comes from dataset size. This was limited because we used human evaluation to label model responses as truthful or untruthful. That is, we have manually confirmed GPT-judge labels on Davinci responses, and extrapolated the system to Ada, Babbage, and Curie. Frankly, the limitations caused by the small size of the dataset were quite evident because the truthfulness detector was often biased towards producing one label (either 1 or 0). We attempted to solve this problem using lower regularization parameters, but this often produced models with lower performances. An ideal solution to this problem would be training the truthfulness detector on a large set of training instances, which is also our future direction.
\bibliography{custom}

\begin{thebibliography}{23}
\expandafter\ifx\csname natexlab\endcsname\relax\def\natexlab#1{#1}\fi

\bibitem[{Albahar(2021)}]{albahar2021hybrid}
Marwan Albahar. 2021.
\newblock A hybrid model for fake news detection: Leveraging news content and
  user comments in fake news.
\newblock \emph{IET Information Security}, 15(2):169--177.

\bibitem[{Bang et~al.(2023)Bang, Cahyawijaya, Lee, Dai, Su, Wilie, Lovenia, Ji,
  Yu, Chung et~al.}]{bang2023multitask}
Yejin Bang, Samuel Cahyawijaya, Nayeon Lee, Wenliang Dai, Dan Su, Bryan Wilie,
  Holy Lovenia, Ziwei Ji, Tiezheng Yu, Willy Chung, et~al. 2023.
\newblock A multitask, multilingual, multimodal evaluation of chatgpt on
  reasoning, hallucination, and interactivity.
\newblock \emph{arXiv preprint arXiv:2302.04023}.

\bibitem[{Brown et~al.(2020)Brown, Mann, Ryder, Subbiah, Kaplan, Dhariwal,
  Neelakantan, Shyam, Sastry, Askell et~al.}]{brown2020language}
Tom Brown, Benjamin Mann, Nick Ryder, Melanie Subbiah, Jared~D Kaplan, Prafulla
  Dhariwal, Arvind Neelakantan, Pranav Shyam, Girish Sastry, Amanda Askell,
  et~al. 2020.
\newblock Language models are few-shot learners.
\newblock \emph{Advances in neural information processing systems},
  33:1877--1901.

\bibitem[{Chen et~al.(2017)Chen, Fisch, Weston, and Bordes}]{chen2017reading}
Danqi Chen, Adam Fisch, Jason Weston, and Antoine Bordes. 2017.
\newblock Reading wikipedia to answer open-domain questions.
\newblock In \emph{Proceedings of the 55th Annual Meeting of the Association
  for Computational Linguistics (Volume 1: Long Papers)}, pages 1870--1879.

\bibitem[{Chen and Yih(2020)}]{chen2020open}
Danqi Chen and Wen-tau Yih. 2020.
\newblock Open-domain question answering.
\newblock In \emph{Proceedings of the 58th annual meeting of the association
  for computational linguistics: tutorial abstracts}, pages 34--37.

\bibitem[{Choudhary and Arora(2021)}]{choudhary2021linguistic}
Anshika Choudhary and Anuja Arora. 2021.
\newblock Linguistic feature based learning model for fake news detection and
  classification.
\newblock \emph{Expert Systems with Applications}, 169:114171.

\bibitem[{Dasgupta et~al.(2022)Dasgupta, Lampinen, Chan, Creswell, Kumaran,
  McClelland, and Hill}]{dasgupta2022language}
Ishita Dasgupta, Andrew~K Lampinen, Stephanie~CY Chan, Antonia Creswell,
  Dharshan Kumaran, James~L McClelland, and Felix Hill. 2022.
\newblock Language models show human-like content effects on reasoning.
\newblock \emph{arXiv preprint arXiv:2207.07051}.

\bibitem[{Etezadi and Shamsfard(2023)}]{etezadi2023state}
Romina Etezadi and Mehrnoush Shamsfard. 2023.
\newblock The state of the art in open domain complex question answering: a
  survey.
\newblock \emph{Applied Intelligence}, 53(4):4124--4144.

\bibitem[{Han et~al.(2021)Han, Zhang, Ding, Gu, Liu, Huo, Qiu, Yao, Zhang,
  Zhang et~al.}]{han2021pre}
Xu~Han, Zhengyan Zhang, Ning Ding, Yuxian Gu, Xiao Liu, Yuqi Huo, Jiezhong Qiu,
  Yuan Yao, Ao~Zhang, Liang Zhang, et~al. 2021.
\newblock Pre-trained models: Past, present and future.
\newblock \emph{AI Open}, 2:225--250.

\bibitem[{Ho et~al.(2020)Ho, Nguyen, Sugawara, and Aizawa}]{ho2020constructing}
Xanh Ho, Anh-Khoa~Duong Nguyen, Saku Sugawara, and Akiko Aizawa. 2020.
\newblock Constructing a multi-hop qa dataset for comprehensive evaluation of
  reasoning steps.
\newblock In \emph{Proceedings of the 28th International Conference on
  Computational Linguistics}, pages 6609--6625.

\bibitem[{Jindal et~al.(2020)Jindal, Sood, Singh, Vatsa, and
  Chakraborty}]{jindal2020newsbag}
Sarthak Jindal, Raghav Sood, Richa Singh, Mayank Vatsa, and Tanmoy Chakraborty.
  2020.
\newblock Newsbag: A multimodal benchmark dataset for fake news detection.
\newblock In \emph{CEUR Workshop Proc.}, volume 2560, pages 138--145.

\bibitem[{Joshi et~al.(2017)Joshi, Choi, Weld, and
  Zettlemoyer}]{joshi2017triviaqa}
Mandar Joshi, Eunsol Choi, Daniel~S Weld, and Luke Zettlemoyer. 2017.
\newblock Triviaqa: A large scale distantly supervised challenge dataset for
  reading comprehension.
\newblock In \emph{Proceedings of the 55th Annual Meeting of the Association
  for Computational Linguistics (Volume 1: Long Papers)}, pages 1601--1611.

\bibitem[{Lin et~al.(2022)Lin, Hilton, and Evans}]{lin2022truthfulqa}
Stephanie Lin, Jacob Hilton, and Owain Evans. 2022.
\newblock Truthfulqa: Measuring how models mimic human falsehoods.
\newblock In \emph{Proceedings of the 60th Annual Meeting of the Association
  for Computational Linguistics (Volume 1: Long Papers)}, pages 3214--3252.

\bibitem[{Manakul et~al.(2023)Manakul, Liusie, and
  Gales}]{manakul2023selfcheckgpt}
Potsawee Manakul, Adian Liusie, and Mark~JF Gales. 2023.
\newblock Selfcheckgpt: Zero-resource black-box hallucination detection for
  generative large language models.
\newblock \emph{arXiv preprint arXiv:2303.08896}.

\bibitem[{Mialon et~al.(2023)Mialon, Dess{\`\i}, Lomeli, Nalmpantis, Pasunuru,
  Raileanu, Rozi{\`e}re, Schick, Dwivedi-Yu, Celikyilmaz
  et~al.}]{mialon2023augmented}
Gr{\'e}goire Mialon, Roberto Dess{\`\i}, Maria Lomeli, Christoforos Nalmpantis,
  Ram Pasunuru, Roberta Raileanu, Baptiste Rozi{\`e}re, Timo Schick, Jane
  Dwivedi-Yu, Asli Celikyilmaz, et~al. 2023.
\newblock Augmented language models: a survey.
\newblock \emph{arXiv preprint arXiv:2302.07842}.

\bibitem[{Mihaylov et~al.(2018)Mihaylov, Clark, Khot, and
  Sabharwal}]{mihaylov2018can}
Todor Mihaylov, Peter Clark, Tushar Khot, and Ashish Sabharwal. 2018.
\newblock Can a suit of armor conduct electricity? a new dataset for open book
  question answering.
\newblock In \emph{Proceedings of the 2018 Conference on Empirical Methods in
  Natural Language Processing}, pages 2381--2391.

\bibitem[{Otegi et~al.(2020)Otegi, Campos, Azkune, Soroa, and
  Agirre}]{otegi-eta-2020-automatic}
Arantxa Otegi, Jon~Ander Campos, Gorka Azkune, Aitor Soroa, and Eneko Agirre.
  2020.
\newblock \href {https://doi.org/10.18653/v1/2020.nlpcovid19-2.15} {Automatic
  evaluation vs. user preference in neural textual {Q}uestion{A}nswering over
  {COVID}-19 scientific literature}.
\newblock In \emph{Proceedings of the 1st Workshop on {NLP} for {COVID}-19
  (Part 2) at {EMNLP} 2020}, Online. Association for Computational Linguistics.

\bibitem[{Shiffrin and Mitchell(2023)}]{shiffrin2023probing}
Richard Shiffrin and Melanie Mitchell. 2023.
\newblock Probing the psychology of ai models.
\newblock \emph{Proceedings of the National Academy of Sciences},
  120(10):e2300963120.

\bibitem[{Weidinger et~al.(2022)Weidinger, Uesato, Rauh, Griffin, Huang,
  Mellor, Glaese, Cheng, Balle, Kasirzadeh et~al.}]{weidinger2022taxonomy}
Laura Weidinger, Jonathan Uesato, Maribeth Rauh, Conor Griffin, Po-Sen Huang,
  John Mellor, Amelia Glaese, Myra Cheng, Borja Balle, Atoosa Kasirzadeh,
  et~al. 2022.
\newblock Taxonomy of risks posed by language models.
\newblock In \emph{2022 ACM Conference on Fairness, Accountability, and
  Transparency}, pages 214--229.

\bibitem[{Welleck et~al.()Welleck, Kulikov, Roller, Dinan, Cho, and
  Weston}]{welleckneural}
Sean Welleck, Ilia Kulikov, Stephen Roller, Emily Dinan, Kyunghyun Cho, and
  Jason Weston.
\newblock Neural text generation with unlikelihood training.
\newblock In \emph{International Conference on Learning Representations}.

\bibitem[{Yang et~al.(2018)Yang, Qi, Zhang, Bengio, Cohen, Salakhutdinov, and
  Manning}]{yang2018hotpotqa}
Zhilin Yang, Peng Qi, Saizheng Zhang, Yoshua Bengio, William Cohen, Ruslan
  Salakhutdinov, and Christopher~D Manning. 2018.
\newblock Hotpotqa: A dataset for diverse, explainable multi-hop question
  answering.
\newblock In \emph{Proceedings of the 2018 Conference on Empirical Methods in
  Natural Language Processing}, pages 2369--2380.

\bibitem[{Zervopoulos et~al.(2022)Zervopoulos, Alvanou, Bezas, Papamichail,
  Maragoudakis, and Kermanidis}]{zervopoulos2022deep}
Alexandros Zervopoulos, Aikaterini~Georgia Alvanou, Konstantinos Bezas,
  Asterios Papamichail, Manolis Maragoudakis, and Katia Kermanidis. 2022.
\newblock Deep learning for fake news detection on twitter regarding the 2019
  hong kong protests.
\newblock \emph{Neural Computing and Applications}, 34(2):969--982.

\bibitem[{Zhao et~al.(2023)Zhao, Zhou, Li, Tang, Wang, Hou, Min, Zhang, Zhang,
  Dong et~al.}]{zhao2023survey}
Wayne~Xin Zhao, Kun Zhou, Junyi Li, Tianyi Tang, Xiaolei Wang, Yupeng Hou,
  Yingqian Min, Beichen Zhang, Junjie Zhang, Zican Dong, et~al. 2023.
\newblock A survey of large language models.
\newblock \emph{arXiv preprint arXiv:2303.18223}.

\end{thebibliography}
\bibliographystyle{acl_natbib}

\end{document}